\documentclass[12pt]{iopart}

\usepackage{cite}
\usepackage{graphicx}
\usepackage{textcomp,nicefrac}
\usepackage{subcaption}
\usepackage{lipsum}
\usepackage{float}
\usepackage{textcomp,nicefrac}
\usepackage[spaces,hyphens]{url}
\usepackage{tabularx}
\usepackage{adjustbox}
\usepackage{siunitx}
\usepackage{color,soul}
\usepackage{esvect}
\usepackage{amsmath} 
\usepackage{multirow}
\usepackage{hyperref}
\usepackage[table]{xcolor}
\usepackage{cleveref}

\newcolumntype{Y}{>{\centering\arraybackslash}X}

\begin{document}

\title[]{rule4ml: An Open-Source Tool for Resource Utilization and Latency Estimation for ML Models on FPGA}

\author{Mohammad Mehdi Rahimifar,
         Hamza Ezzaoui Rahali,  
	   and Audrey C. Therrien}

         \vspace{2mm}

\address{Interdisciplinary Institute for Technological Innovation - 3IT, Université de Sherbrooke, Sherbrooke (Québec), Canada}
\vspace{1mm}

\vspace{1mm}
\ead{Mohammad.Mehdi.Rahimifar@Usherbrooke.ca}
\begin{indented}
\item[]July 2024
\end{indented}

\begin{abstract}\\

Implementing Machine Learning (ML) models on Field-Programmable Gate Arrays (FPGAs) is becoming increasingly popular across various domains as a low-latency and low-power solution that helps manage large data rates generated by continuously improving detectors. However, developing ML models for FPGAs is time-consuming, as optimization requires synthesis to evaluate FPGA area and latency, making the process slow and repetitive. This paper introduces a novel method to predict the resource utilization and inference latency of Neural Networks (NNs) before their synthesis and implementation on FPGA. We leverage HLS4ML, a tool-flow that helps translate NNs into high-level synthesis (HLS) code, to synthesize a diverse dataset of NN architectures and train resource utilization and inference latency predictors. While HLS4ML requires full synthesis to obtain resource and latency insights, our method uses trained regression models for immediate pre-synthesis predictions. The prediction models estimate the usage of Block RAM (BRAM), Digital Signal Processors (DSP), Flip-Flops (FF), and Look-Up Tables (LUT), as well as the inference clock cycles. The predictors were evaluated on both synthetic and existing benchmark architectures and demonstrated high accuracy with R\textsuperscript{2} scores ranging between 0.8 and 0.98 on the validation set and sMAPE values between 10\% and 30\%. Overall, our approach provides valuable preliminary insights, enabling users to quickly assess the feasibility and efficiency of NNs on FPGAs, accelerating the development and deployment processes. The open-source repository can be found at \href{https://github.com/IMPETUS-UdeS/rule4ml}{https://github.com/IMPETUS-UdeS/rule4ml}, while the datasets are publicly available at \href{https://borealisdata.ca/dataverse/rule4ml}{https://borealisdata.ca/dataverse/rule4ml}.

\end{abstract}


\clearpage

\section{Introduction}

Neural Networks (NNs) are a subset of Machine Learning (ML) models, popularized due to their ability to model complex patterns and relationships in data. These algorithms are transforming several fields such as healthcare, finance, manufacturing, and entertainment by enabling predictive analytics, automating processes, and enhancing decision-making~\cite{rahali2024efficient, li2017survey}. In particular, the surge in data generated by advanced detectors over the past decade has created a need for data compression at the source, a task at which NNs have proven highly effective~\cite{khoda2023ultra,therrien2022potential}. However, the high data rates from instrumentation require processing units with minimal latency to ensure real-time performance from NN implementation~\cite{parra2018systematic}.\\

Compared to common processing units such as Central Processing Units (CPU) and Graphics Processing Units (GPU), Field Programmable Gate Arrays (FPGAs)  provide a large number of Input/Output (I/O) ports, and their architectures are designed to fully exploit pipelining and parallelism, enabling continuous and concurrent computing~\cite{nurvitadhi2016accelerating}. The implementation of FPGA-based Edge Machine Learning (EdgeML) is an active area of research, particularly targeting instrumentation applications that require minimal latency.  For instance, a fast inference of boosted decision trees on FPGAs near the sensor is used in particle physics applications~\cite{summers2020fast}. Ultra-low-latency and low-power neural networks with convolutional layers are also employed on FPGAs for various instrumentation applications~\cite{aarrestad2021fast}. Additionally, FPGA-embedded systems are being developed for ML-based tracking and triggering in electron–ion collider experiments~\cite{hong2021nanosecond}.\\

Developing NNs on FPGA using traditional Hardware Description Language (HDL) requires significant time and effort, as well as a high understanding of hardware architecture~\cite{lahti2018we}. Accordingly, some tool-flows have been developed to ease the implementation of NNs on FPGA. The tool-flows translate NNs implemented using common ML libraries into High-Level Synthesis (HLS) code, enabling FPGA implementation with minimum hardware knowledge. In our previous review~\cite{rahimifar2023exploring}, we determined that HLS4ML~\cite{fahim2021hls4ml} stands out as the current best option, considering its wide support for different NN architectures, and its ability to adapt to the application requirements and achieve low power and low latency, both essential for EdgeML solutions and instrumentation applications. However, in its current state, HLS4ML cannot indicate whether an NN will fit the targeted board before synthesis. This is a significant limitation, as larger NNs may require several hours to synthesize, and the resource utilization report is only available after the synthesis phase. The resource utilization report details the usage of Block RAM (BRAM), Digital Signal Processors (DSP), Flip-Flops (FF), and Look-Up Tables (LUT), revealing whether the network implementation exceeds the capacities of a board. Additionally, HLS4ML currently lacks a latency approximation for the NNs before synthesis, which is crucial for projects with strict timing constraints. In this context, latency refers to the number of clock cycles required for a single inference of an NN on FPGA. Moreover, high resource utilization on FPGA can also cause timing issues during the implementation phase, resulting in implementation failure~\cite{cong2022fpga}. In this case, a time-consuming re-synthesis is necessary after compressing the network and adjusting the synthesis parameters, taking away from the otherwise fast development cycle provided by HLS4ML.\\

This paper investigates the use of ML models to predict resource utilization and inference latency for NNs before synthesis. We seek to accelerate the prototyping and deployment of ML solutions on FPGAs, providing HLS4ML users accurate, early, and immediate insight into the behavior of their NNs across various boards and configurations. The prediction can also help identify possible optimizations in advance, substantially reducing the time needed to achieve an ideal design. There has not been any previously published research specifically addressing the prediction of FPGA resources and latency for NNs, highlighting the originality of our approach in this study.\\

The rest of the paper is organized as follows: Section II explains the HLS4ML architecture and how its configurations significantly affect an NN's resource utilization and inference latency. Section III describes the data generation process, the prediction approach, and the ML models trained to approximate FPGA resource utilization and latency. The prediction results are presented in Section IV. Finally, we discuss the results and conclude the paper in sections V and VI respectively.\\

\section{Background}

\subsection{HLS4ML}

HLS4ML stands out among several NN-to-FPGA implementation tool-flows for its exceptional performance in key areas: minimal latency, efficient resource utilization, user-friendliness, and configuration flexibility~\cite{rahimifar2023exploring}. The effectiveness of HLS4ML has been demonstrated in various research projects implementing diverse NNs on FPGAs.  For example, a real-time semantic segmentation on FPGAs for autonomous vehicles with minimal latency was achieved using HLS4ML~\cite{ghielmetti2022real}. Ultra-low latency Recurrent Neural Network (RNN) inference on FPGAs for physics applications has also been studied~\cite{khoda2023ultra}. Additionally, HLS4ML was used to explore efficient compression algorithms at the edge for real-time data acquisition in a billion-pixel X-ray camera in one of our previous works~\cite{rahali2024efficient}.\\

\begin{figure}[t]
\centering
\includegraphics[width=0.95\textwidth]{ 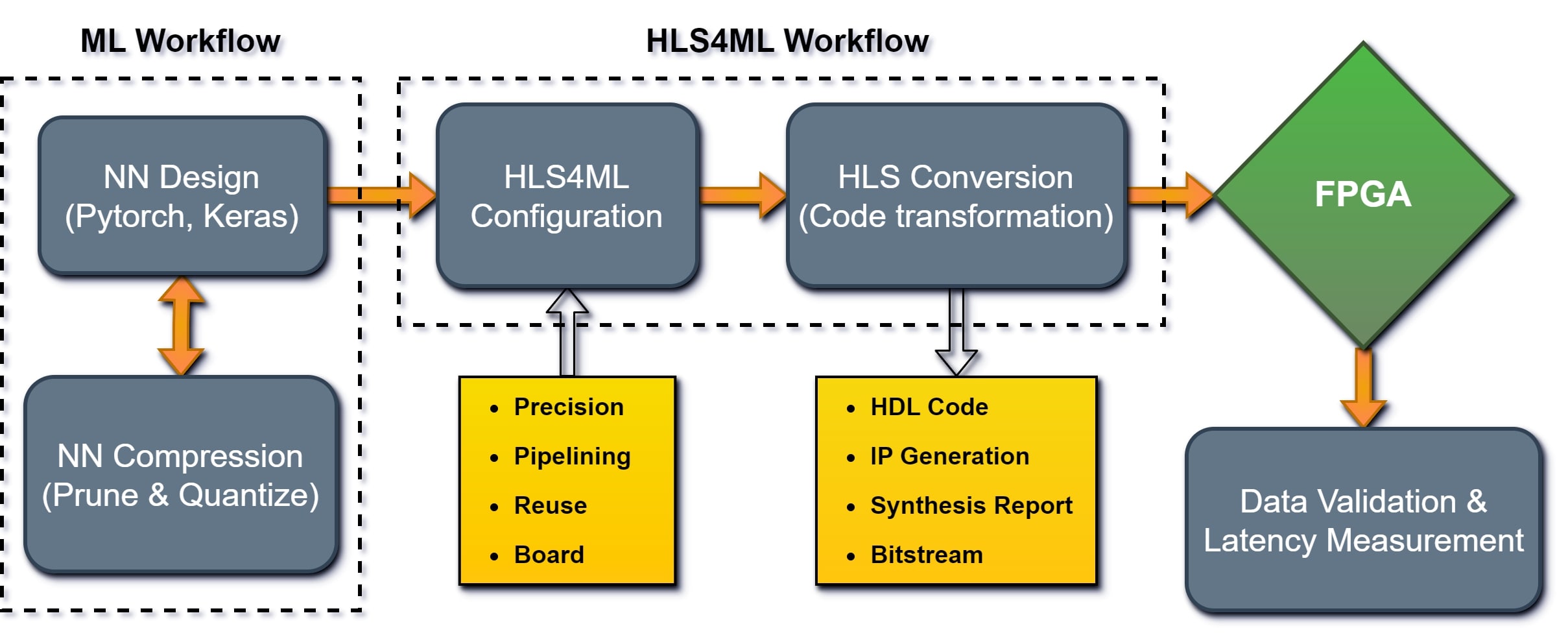}
\caption{HLS4ML workflow.}
\label{fig:hls4ml_workflow}
\end{figure}

Figure~\ref{fig:hls4ml_workflow} illustrates how HLS4ML translates an NN from standard Python code into HLS code ready for FPGA implementation. The first step involves designing an NN architecture for the targeted task. This can be done using \textit{Keras} or \textit{PyTorch} libraries, possibly using the Open Neural Network Exchange (ONNX) format. Tools such as \textit{QKeras} help in network quantization for FPGA resource efficiency. Subsequently, users can fine-tune HLS4ML settings like precision, reuse, and choose the target board. Finally, HLS4ML translates the code to HLS, producing the HDL code together with the Vivado IP. For supported boards~\cite{hls4mlgithub}, HLS4ML can generate the complete block design and bitstream for direct FPGA programming without requiring additional user intervention.\\

The next section explains how we have used HLS4ML to synthesize a large dataset of synthetic NNs with various architectures and configurations. The data was fed to ML models to train accurate predictors capable of predicting resource usage and latency, requiring only the NN architecture as input.

\section{Methodology} 

\subsection{Data generation}
\par{

To train regression models capable of predicting both the resource utilization and the inference latency of an HLS4ML-synthesized NN, we generated and synthesized a diverse collection of NN architectures. For accurate predictions, the training dataset must encompass a wide range of architecture-related parameters, including various input and output sizes, different numbers of layers and neurons, along with various types of layers and operations: matrix multiplication, non-linear activations, skip connections, batch normalization, etc. However, for the current iteration, we choose to exclude computationally taxing operations such as convolution, speeding up the synthesis and data generation processes.\\
}

\begin{table}[t]
\scriptsize
\caption{Data generation parameters and their ranges.}
\renewcommand{\arraystretch}{1.3}
\centering
\begin{tabular}{@{}c@{}|@{}c@{}|@{}c@{}}
    \begin{tabularx}{0.195\textwidth}{|Y|} \hline
        \\ \hline
        \\
        \\
        \textbf{Neural Network} \\
        \textbf{Parameters} \\
        \\ \hline \hline
        \\
        \textbf{HLS4ML} \\
        \textbf{Parameters} \\
        \\ \hline
    \end{tabularx}
    \begin{tabular}{|c|c|c} \hline
        \textbf{Parameter} & \textbf{Value range} & \textbf{Range type} \\ \hline
        Input size & [16, 1024] & Powers of 2\\
        Layer count & [2, 20] & Increments of 1 \\
        Neuron count & [2, 4096] & Powers of 2 \\
        Output size & [1, 200] & Increments of 1 \\
        Activation function & ReLU, Tanh, Sigmoid, Softmax & - \\ \hline \hline
        Precision/Bit width & 2, 8, 16 & - \\
        Reuse Factor & [1, 64] & Powers of 2 \\
        Target Board & Pynq-z2, ZCU102, Alveo-u200 & - \\
        Strategy & Latency, Resource & - \\ \hline
    \end{tabular}
\end{tabular}
\label{table:data-gen-params}
\vspace{-4mm}
\end{table}

The parameters of a synthetic NN are randomly selected from their respective ranges listed in Table~\ref{table:data-gen-params}. A few additional HLS4ML-related parameters are required for the synthesis:

\begin{itemize}
    \item[$\bullet$] {Precision}: Specifies the total number of bits used to represent each of the inputs, outputs, weights, and biases of layers.
    \item[$\bullet$] {Reuse Factor}: Determines the number of times an FPGA multiplier is reused, controlling the level of pipelining, and directly impacting the latency and resources.
    \item[$\bullet$] {Strategy}: Chooses whether to prioritize resource or latency optimization.
    \item[$\bullet$] {Board}: Selects the target board, for which the synthesis report is generated.
\end{itemize}

The value ranges and their increments were selected to encapsulate a wide variety of NNs, targeting different tasks (prediction, classification, clustering, etc.), all while making sure the architectures remain within reasonable bounds, both in terms of duration and feasibility of the synthesis. While HLS4ML's precision and strategy parameters can be configured on a per-layer basis, we chose to assign a global value across all layers of the generated network, limiting the otherwise massive number of possible combinations. The reuse factor, however, can still change from one layer to another since its maximum value is bounded by a layer's input and output sizes.\\

In addition to the primary parameters listed in Table~\ref{table:data-gen-params}, probability functions dictate the use of certain layers, such as skip connections and batch normalization. These probabilities are based on the network's size and complexity, enhancing the dataset's representation of real-world architectures. The dataset contains more than 15,000 synthetic NNs, generated within approximately 28,800 CPU-core hours. The synthesis ran in parallel on 2 x Intel Xeon Processor E5-2660 v2 and 2 x Intel Xeon Processor E5-2680 v4. The project repository\footnote[1]{Repository: \href{https://github.com/IMPETUS-UdeS/rule4ml}{https://github.com/IMPETUS-UdeS/rule4ml}} includes the code used for data generation, while the datasets\footnote[2]{Datasets: \href{https://borealisdata.ca/dataverse/rule4ml}{https://borealisdata.ca/dataverse/rule4ml}} can be accessed separately.

\subsection{Feature engineering}

As networks are generally structured in a sequence of layers, all architecture-related data is inherently layer-dependent, where each layer can have its own set of parameters. While some parameters are universally applicable, such as the layer type (e.g., fully connected, activation, batch normalization) and its input and output sizes, others such as the number of neurons, bias, and trainable parameters are exclusive to certain layers within the architecture. Consequently, there are two main approaches for training the predictors: either choose a regression model architecture capable of handling the sequential nature of the inputs or convert the raw data into a format compatible with a wider range of architectures.\\

Some ML models are inherently designed to operate on sequential data: architectures such as RNNs and Transformers can capture local and global dependencies within sequences~\cite{hochreiter1997-lstm, vaswani2017-transformer}. At first glance, these models might appear well-suited to the sequential nature of our data. However, training RNNs can be challenging due to vanishing and exploding gradients~\cite{pascanu2013-rnn}, while Transformers are prone to be overly complex and fail to achieve state-of-the-art performance on some tasks~\cite{zeng2023-transformer}. Alternatively, extracting and creating meaningful network-level features from the raw sequential data enables the use of regression ML models like decision trees, random forests, or Multi-Layer Perceptron (MLP).  \\

While these simpler models are easier to train, tune, and interpret, ultimately our goal is to achieve the highest accuracy for the predictions. The evaluation of the models will be the object of the next section, using both sequential raw inputs and extracted features. The rest of this section will focus on feature engineering, extracting network-level features from the original layer-wise data.\\

The raw features are distilled into more meaningful engineered features mainly through aggregation and statistical averages. Additional features are calculated from the number of fixed-point operations required for a layer's implementation in HLS4ML. The fixed-point operations include addition, multiplication, logical operations, and lookup operations, i.e. retrieving a value from a pre-computed table using an index. Engineered features, extracted from the raw data for each network in the dataset, are listed below:

\begin{itemize}
    \item \textbf{Aggregated Features}: The total counts of dense layers, batch normalization layers, skip layers, dropout layers, and the count of each type of activation.
    \item \textbf{Statistical Features}: The average number of parameters, inputs, outputs, and the average reuse factor of fully connected layers specifically, as they are expected to be the most computationally intensive.
    \item \textbf{Fixed-point Operations}: The single counts of addition, multiplication, logical, and lookup operations used across the entire architecture, multiplied by the HLS4ML precision.
\end{itemize}

Meanwhile, the dependent resource variables (BRAM, DSP, FF, and LUT), having unbounded and considerably different ranges of values, are scaled down to a normalized interval representing the NN's utilization percentage of the target board's available resources. The lower limit of the range is 0\%, while the upper bound is set to 200\%, considering possible inaccuracies within the synthesis reports. We expect NNs using more than 200\% of any resource to fail the implementation phase, regardless of potential post-synthesis optimizations. Therefore, we consider prediction accuracy irrelevant for those values, limiting them to the set upper bound.\\

\begin{figure}[b]
\centering
\includegraphics[width=0.95\textwidth]{ 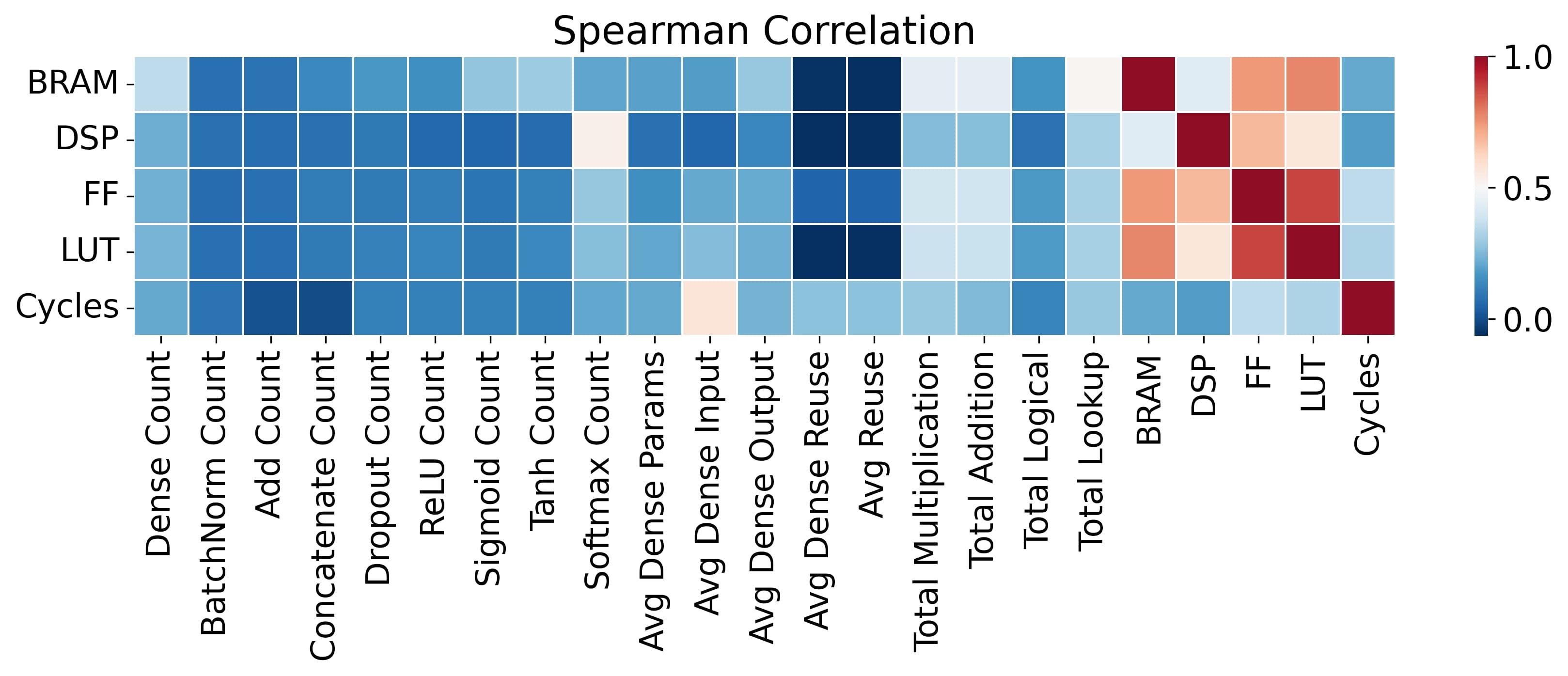}
\caption{Correlation matrix showing the interdependence between variables.}
\label{fig:corr-matrix}
\end{figure}

Furthermore, we calculated the Spearman correlation~\cite{schober2018} to study dependencies between features and prediction variables. The correlation matrix depicted in Figure~\ref{fig:corr-matrix} illustrates the relationship between the engineered features and the resources and latency values. For the generated dataset, the resources and latency (cycles) primarily correlate with features extracted from fully connected layers, plus the summed numbers of fixed-point operations. As expected, the reuse factor appears to have a higher positive correlation with clock cycles and a negative correlation with resources. The low correlation between resource utilization and reuse factor can be attributed to our dataset's composition. The dataset is dominated by larger boards (ZCU102 and Alveo-u200) and lower precisions (2 and 8-bit). These factors inherently lead to lower resource consumption, reducing the impact of reuse on overall resource utilization. Other unexpected trends are also noticeable: the softmax layer seems highly correlated with the DSPs despite its HLS4ML implementation mainly relying on lookup tables.

\subsection{Model selection}

To evaluate the approaches, we trained regression models both with and without the engineered features. Specifically, we experimented with Long Short-Term Memory (LSTM) and Transformers~\cite{yu2019review}, Random Forests~\cite{breiman2001random}, Gradient Boosted Trees~\cite{ke2017lightgbm}, and MLP models. Feature engineering was applied to extract meaningful inputs for simpler models, while the sequential ones (LSTM and Transformer) were trained on the raw, layer-based data. The results of the preliminary training are summarized in Table~\ref{table:ML-comparison}, highlighting the performance differences between the different models.\\

\begin{table}[b]
\footnotesize
\caption{Comparing performance of ML models on the training set using R\textsuperscript{2} score.}
\renewcommand{\arraystretch}{1.8}
\setlength\tabcolsep{7pt}
\centering
\begin{tabular}{c|ccccc}
\hline
\multirow{2}{*}{\textbf{ML Model}} & \multicolumn{5}{c}{\textbf{R\textsuperscript{2} Score~\cite{chicco2021}}} \\
\cline{2-6} &
\textbf{BRAM} & \textbf{DSP} & \textbf{FF} & \textbf{LUT} & \textbf{Cycles} \\
\hline
LSTM & 0.81 & \cellcolor{red!70}0.03 & \cellcolor{red!70}0.48 & \cellcolor{red!70}0.74 & \cellcolor{red!70}0.80 \\
Transformer & \cellcolor{red!70}0.78 & \cellcolor{green!75}0.93 & \cellcolor{green!75}0.85 & \cellcolor{red!70}0.74 & \cellcolor{green!75}0.88 \\
Random Forest & 0.89 & 0.88 & 0.79 & 0.83 & 0.86 \\
Gradient Boosted Trees & 0.84 & 0.86 & 0.77 & \cellcolor{green!75}0.86 & \cellcolor{green!75}0.88 \\
MLP & \cellcolor{green!75}0.92 & 0.85 & 0.72 & 0.85 & 0.87 \\ \hline
\end{tabular}
\vspace{2mm}
\label{table:ML-comparison}
\end{table}

Table~\ref{table:ML-comparison} shows the R\textsuperscript{2} score for each dependent variable, where higher values indicate better predictive performance. The table highlights the best predictor for each variable in green and the worst in red. Accordingly, the LSTM scores the worst for all variables except BRAM. The Transformer performs the best for DSP, FF, and clock cycle predictions on the validation set, but also the worst for BRAM and ties with the LSTM for the lowest LUT score. Meanwhile, the models using the extracted, network-level features perform consistently well. Boosted trees achieved the highest LUT score and shared the best cycle score with the Transformer. The MLP performs the best for BRAM and comes very close to the LUT and cycles scores of boosted trees. Considering the initial results, the simpler models using extracted features are largely preferred to the LSTM and Transformer. Additionally, we decided to use one type of architecture across variables to keep the training, fine-tuning, and testing processes uniform and straightforward. The remainder of this work will focus on MLPs as the model architecture of predictors, which show consistent performance and are easier to fine-tune.\\

We used the Hyperband~\cite{li2018hyperband} and Bayesian optimization~\cite{snoek2012practical} algorithms to fine-tune the hyperparameters, the number, and size of layers, as well as activation functions of the MLPs. The algorithms optimized a board-relative mean absolute error (MAE) between the predictions and actual values of each resource variable and a standard MAE for the clock cycles. Compared to Random and Grid search~\cite{liashchynskyi2019grid} algorithms, the Hyperband and Bayesian optimization demonstrate efficiency with quicker convergence and early stopping, significantly reducing the time needed for fine-tuning. To avoid issues with the scale and distribution differences between the dependent variables, we opted to train separate regression models for each resource and an additional one for latency prediction. The use of multiple models also led to lower MAEs compared to a single model predicting all variables at once.\\

\begin{figure}[t]
    \centering
    \includegraphics[width=0.95\linewidth]{ 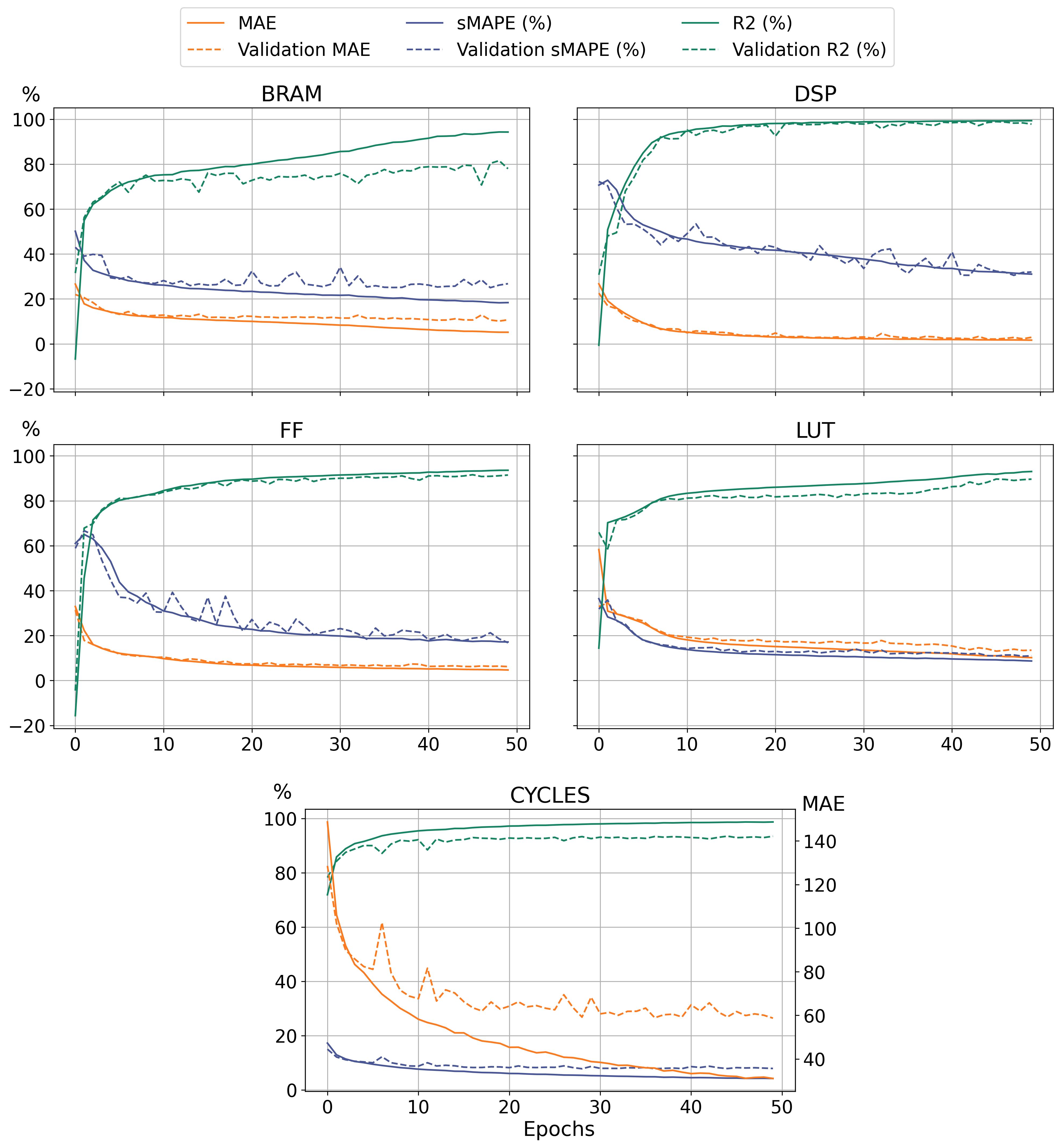}
    \caption{Training and validation loss and metrics. The MAE loss for BRAM, DSP, FF, and LUT is a percentage relative to the board's available resources. Standard MAE is used for clock cycles.}
    \label{fig:train-history}
\end{figure}

Table~\ref{NN_setups} presents the models' architectures alongside their fine-tuned hyperparameters. Ordinal encoding~\cite{potdar2017} is applied to categorical inputs (board, strategy, etc.), which are then processed through trainable embedding layers. Numerical inputs are fed into a dense block (Dense Block 1), then concatenated with embedding layers outputs, and finally passed through a second dense block (Dense Block 2). Each model is trained over 50 epochs using the ADAM optimizer~\cite{kingma2014Adam}, with ReLU activations applied after dense layers. The MAE loss, sMAPE~\cite{makridakis1993} and R\textsuperscript{2} scores of the models during training and validation are presented in Figure~\ref{fig:train-history}. The weights and configurations of the best-performing models for each resource and latency variable are saved for further predictions.

\begin{table}[b]
\footnotesize
\caption{Fine-tuned Training Configurations for Resource Utilization and Latency Prediction Models.}
\renewcommand{\arraystretch}{1.7}
\setlength\tabcolsep{4pt}
\centering
\begin{tabular}{lp{1.05cm}p{1.55cm}cc}
\hline
\textbf{Model} & \textbf{Batch} \newline \textbf{Size} & \textbf{Learning} \newline \textbf{Rate} & \textbf{Dense Block 1} & \textbf{Dense Block 2} \\ \hline
BRAM & 64 & 10\textsuperscript{-4} & $\xrightarrow[]{32}\  \xrightarrow[]{16}\ \xrightarrow[]{32}\ $ & $\xrightarrow[]{256}\  \xrightarrow[]{256}\ \xrightarrow[]{256}\ \xrightarrow[]{64}\ \xrightarrow[]{32}\ \xrightarrow[]{64}\
\xrightarrow[]{64}\  $ \\

DSP & 32 & 10\textsuperscript{-4} & $\xrightarrow[]{64}\  \xrightarrow[]{32}\ \xrightarrow[]{32}\ $ & $\xrightarrow[\text{}]{256}\ \xrightarrow[]{16}\ \xrightarrow[\text{}]{32}\ \xrightarrow[\text{}]{32}\ 
\xrightarrow[\text{}]{64}\ $ \\

FF & 64 & 10\textsuperscript{-4} & $\xrightarrow[]{64}\  \xrightarrow[]{16}\ \xrightarrow[]{32}\ $ & $\xrightarrow[\text{}]{64}\ \xrightarrow[\text{}]{128}\ \xrightarrow[\text{}]{64}\  \xrightarrow[\text{}]{256}\
\xrightarrow[\text{}]{32}\ $ \\

LUT & 32 & 10\textsuperscript{-4} & $\xrightarrow[]{64}\  \xrightarrow[]{16}\ \xrightarrow[]{32}\ \xrightarrow[]{32}\ $ & $\xrightarrow[\text{}]{64}\ \xrightarrow[\text{}]{128}\ \xrightarrow[\text{}]{128}\ \xrightarrow[\text{}]{64}\  $ \\

Cycles & 64 & 10\textsuperscript{-3} & $\xrightarrow[]{32}\  \xrightarrow[]{16}\ \xrightarrow[]{64}\ $ & $\xrightarrow[\text{}]{256}\ \xrightarrow[\text{}]{32}\ \xrightarrow[\text{}]{32}\ \xrightarrow[\text{}]{32}\ \xrightarrow[\text{}]{256}\ \xrightarrow[\text{}]{128}\ \xrightarrow[\text{}]{128}\ \xrightarrow[\text{}]{32}\ \xrightarrow[\text{}]{16}\ \xrightarrow[\text{}]{16}\  \xrightarrow[\text{}]{64}\ $ \\ \hline
\end{tabular}
\vspace{2mm}
\label{NN_setups}
\end{table}

\subsection{Model validation}

We validate the prediction models' performance in a two-step process. First, a test set is randomly split from the synthetic dataset. This set is used to evaluate predictions on a wide range of NN architectures. Second, we prepared a collection of NNs, mostly focused on instrumentation applications for realistic "benchmark" data. Table~\ref{table:architecture} is an overview of all benchmark architectures.\\

\begin{table}[t]
\footnotesize
\caption{Overview of the benchmark's neural networks.}
\renewcommand{\arraystretch}{1.8}
\centering
\begin{tabular}{c|ccc}
\hline
    Benchmark & Size (Params.)  & Input Size & Architecture \\ \hline
    JET~\cite{fahim2021hls4ml} & 2,821 & 16 & $\xrightarrow[\text{ReLU}]{32}\ \xrightarrow[\text{ReLU}]{32}\ \xrightarrow[\text{ReLU}]{32}\ \xrightarrow[\text{ReLU}]{32}\ \xrightarrow[\text{Softmax}]{5}$ \\
    Top quarks~\cite{duarte2018fast} & 385 & 10 & $\xrightarrow[\text{ReLU}]{32}\ \xrightarrow[\text{Sigmoid}]{1}$ \\
    Anomaly~\cite{borras2022open} & 2,864 & 128 & $\xrightarrow[\text{ReLU}]{8}\ \xrightarrow[\text{ReLU}]{4}\ \xrightarrow[\text{ReLU}]{128}\ \xrightarrow[\text{ReLU}]{4}\ \xrightarrow[\text{Softmax}]{128}$ \\
    BiPC~\cite{rahali2024efficient} & 7,776 & 36 & $\xrightarrow[\text{ReLU}]{36}\ \xrightarrow[\text{ReLU}]{36}\ \xrightarrow[\text{ReLU}]{36}\ \xrightarrow[\text{ReLU}]{36}\ \xrightarrow[\text{ReLU}]{36}$ \\
    CookieBox~\cite{gouin2022data} & 3,433 & 512 & $\xrightarrow[\text{ReLU}]{4}\ \xrightarrow[\text{ReLU}]{32}\ \xrightarrow[\text{ReLU}]{32}\ \xrightarrow[\text{Softmax}]{5}$ \\
    MNIST~\cite{MNIST} & 12,730 & 784 & $\xrightarrow[\text{ReLU}]{16}\ \xrightarrow[\text{Softmax}]{10}$ \\
    Automlp~\cite{chen2023automlp} & 534 & 7 & $\xrightarrow[\text{ReLU}]{12}\ \xrightarrow[\text{ReLU}]{16}\ \xrightarrow[\text{ReLU}]{12}\ \xrightarrow[\text{Softmax}]{2}$ \\
    Particle Tracking~\cite{abidi2022charged} & 2,691 & 14 & $\xrightarrow[\text{ReLU}]{32}\ \xrightarrow[\text{ReLU}]{32}\ \xrightarrow[\text{ReLU}]{32}\ \xrightarrow[\text{Softmax}]{3}$ \\[-0.5cm] \\ \hline \hline
    Custom 1 & 5,610 & 16 & $\xrightarrow[\text{ReLU}]{64}\ \xrightarrow[\text{ReLU}]{32}\ \xrightarrow[\text{ReLU}]{32}\ \xrightarrow[\text{ReLU}]{32}\
    \xrightarrow[\text{Softmax}]{10}$ \\
    Custom 2 & 11,074 & 128 & $\xrightarrow[\text{ReLU}]{16}\ \xrightarrow[\text{ReLU}]{64}\ \xrightarrow[\text{ReLU}]{32}\ \xrightarrow[\text{ReLU}]{64}\
    \xrightarrow[\text{ReLU}]{32}\
    \xrightarrow[\text{Softmax}]{50}$ \\
    Custom 3 & 7,274 & 64 & $\xrightarrow[\text{ReLU}]{32}\ \xrightarrow[\text{ReLU}]{32}\ \xrightarrow[\text{ReLU}]{32}\ \xrightarrow[\text{Softmax}]{10}$ \\[-0.5cm] \\ \hline
\end{tabular}
\label{table:architecture}
\vspace{-4mm}
\end{table}

The benchmark set includes "JET"~\cite{fahim2021hls4ml} for particle jet classification and "Top Quarks"~\cite{duarte2018fast} for top quark event classification. The "Anomaly"~\cite{borras2022open} network is used to separate normal and abnormal signals in audio files, while "BiPC"~\cite{rahali2024efficient} and "CookieBox"~\cite{gouin2022data} are architectures for handling real-time data acquisition in a billion-pixel camera and CookieBox setups, respectively. We also included "MNIST"~\cite{MNIST} for digit classification and "Automlp"~\cite{chen2023automlp} for MLP acceleration on FPGAs. Lastly, "particle tracking" refers to an architecture deployed on FPGA for charged particle tracking~\cite{abidi2022charged}. In addition to these literature-based architectures, we introduced three other custom networks designed to test specific layers, such as batch normalization and skip connections with many layers and parameters.\\

To fully test the trained predictors on the benchmark networks, we synthesized them utilizing a large combination of HLS4ML parameters: board types (ZCU102, Pynq-Z2), strategies (Latency, Resource), precisions (2-bit, 8-bit, and 16-bit widths), and reuse factors (1, 2, 4, 8, 16, 32, 64). With 11 benchmark NNs listed in Table~\ref{table:architecture}, this results in $924$ combinations. We denote this as the synthesis ground truth of the benchmark. We employ our trained prediction models to estimate the resource utilization and latency for all combinations and compare them side-by-side with the ground truth, evaluating the predictive performance.

\section{Results}

\begin{figure}[t]
\centering
\includegraphics[width=0.9\textwidth]{ 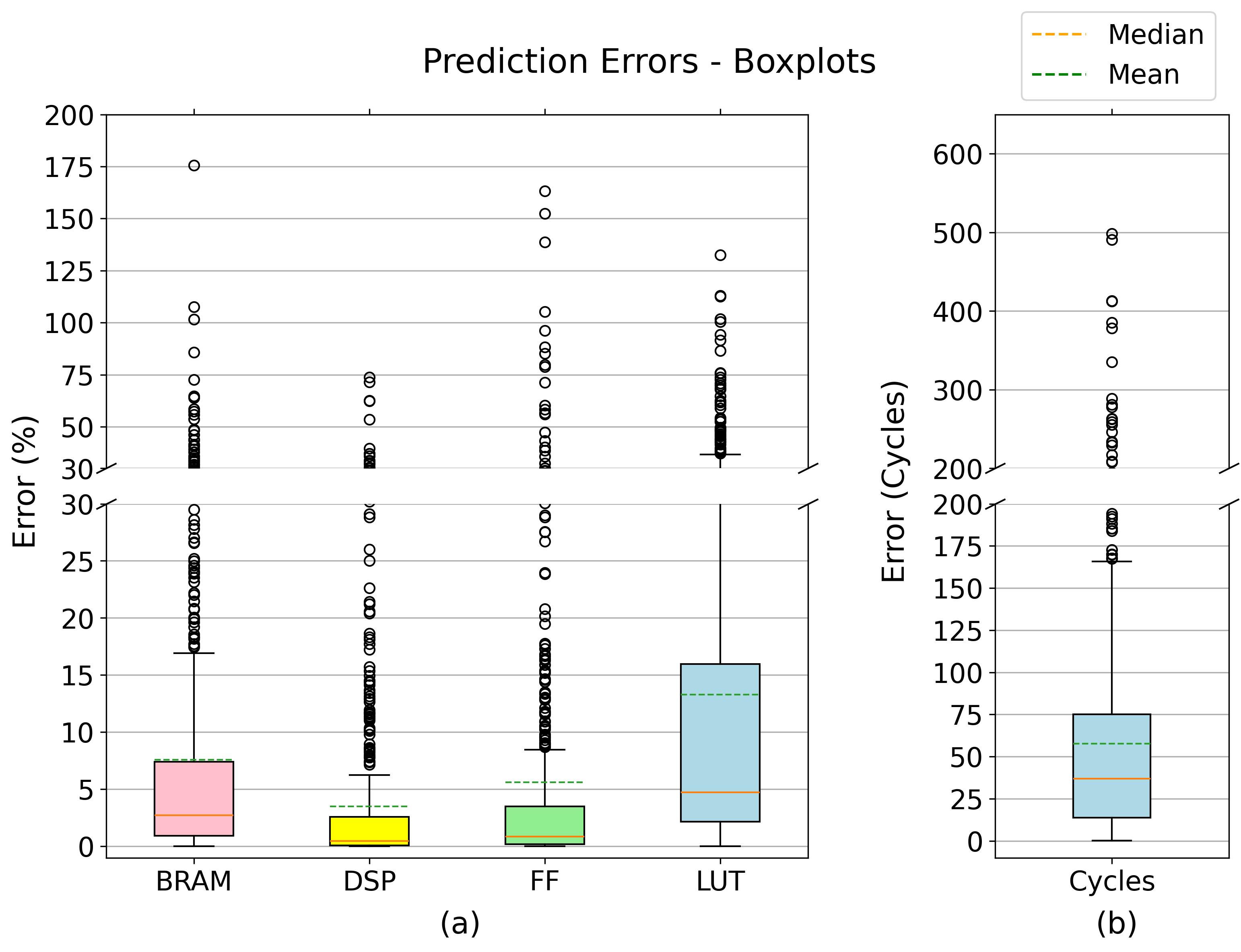}
\caption{Prediction errors illustrated as box plots: (a) resource prediction errors and (b) latency prediction errors (b). The y-axis is broken to better display both the boxes and the outliers, ensuring the scale accommodates accordingly.}
\label{fig:boxplot-merged}
\end{figure}

To assess the models' predictive performance, an initial evaluation is conducted on a test set of 500\texttt{+} models split from the synthetic data, generated according to Table~\ref{table:data-gen-params}. The results are depicted in Figure~\ref{fig:boxplot-merged} using box plots, with the resource predictions on the left~(a) and latency prediction on the right~(b). As illustrated in~(a), the median error for all resources is very low, suggesting that the predictions are generally accurate. The mean prediction errors for LUT is around 10\%, while the errors for BRAM, DSP, and FF are even lower. The Interquartile Range (IQR), the range between the first (Q1) and third quartiles (Q3), is also tight, indicating that most predictions are closely packed around the median. Additionally, the outliers for all resources seem mostly skewed toward lower values, closer to the Q3 + 1.5x IQR range. \\

The right box plot shows the prediction errors for latency, measured in the number of cycles. Similar to the resources box plots, the median and mean are low and within the acceptable range. However, the IQR is large since the errors are not measured in percentages but rather using absolute values, resulting in higher prediction variability. There are fewer outliers compared to resource predictions, with a maximum error of around 500 cycles. This indicates that while most latency predictions are accurate, there are a few significant deviations. Additionally, outliers are also skewed toward lower values and closer to the Q3 + 1.5xIQR area. \\

The second experiment aims to evaluate the fine-tuned regression models on the selected benchmark architectures (Table~\ref{table:architecture}). This validates the model predictions on applied NNs with real use cases. Additionally, we assess whether the predictions follow the trends of the ground truth when adjusting HLS4ML parameters, such as precision and reuse factor.\\

\begin{table}[t]
\footnotesize
\caption{Comparing ground truth (G) and prediction (P) resource utilization percentage for different benchmarks on ZCU102.}
\renewcommand{\arraystretch}{1.3}
\setlength\tabcolsep{4pt}
\begin{center}
\begin{tabular}{c|cc|cc|cc|cc|cc|cc|cc|cc}
\hline
\multirow{3}{*}{Benchmark} & \multicolumn{8}{c|}{Latency} & \multicolumn{8}{c}{Resource} \\ 
\cline{2-17} 
                               & \multicolumn{2}{c|}{BRAM}   & \multicolumn{2}{c|}{DSP}    & \multicolumn{2}{c|}{FF} & \multicolumn{2}{c|}{LUT}    & \multicolumn{2}{c|}{BRAM}   & \multicolumn{2}{c|}{DSP}   & \multicolumn{2}{c|}{FF}   & \multicolumn{2}{c}{LUT}   \\  
\cline{2-17} 
                               & \multicolumn{1}{c}{G\textsuperscript{1}} & \multicolumn{1}{c|}{P\textsuperscript{2}} & \multicolumn{1}{c}{G} & \multicolumn{1}{c|}{P}& \multicolumn{1}{c}{G} & \multicolumn{1}{c|}{P}& \multicolumn{1}{c}{G} & \multicolumn{1}{c|}{P}& \multicolumn{1}{c}{G} & \multicolumn{1}{c|}{P}& \multicolumn{1}{c}{G} & \multicolumn{1}{c|}{P}& \multicolumn{1}{c}{G} & \multicolumn{1}{c|}{P}& \multicolumn{1}{c}{G} & \multicolumn{1}{c}{P}  \\ 
\hline \hline
JET~\cite{fahim2021hls4ml} 
& 0&1 & 0&0 & 1&1 & 25&25 & 1&1 & 0&0 & 1&1 & 9&15 \\

Top Quarks~\cite{duarte2018fast} 
& 0&1 & 0&0 & 0&1 & 6&4 & 0&1 & 0&0 & 0&0 & 7&7 \\

Anomaly~\cite{borras2022open} 
& 0&0 & 5&4 & 11&11 & 76&64 & 2&1 & 5&4 & 12&14 & 90&49 \\

BiPC~\cite{rahali2024efficient} & 0&0 & 0&0 & 1&1 & 68&50 & 3&1 & 0&0 & 3&2 & 24&29 \\

CookieBox~\cite{gouin2022data} 
& 0&1 & 0&0 & 1&1 & 40&38 & 2&1 & 0&0 & 3&4 & 92&81 \\

MNIST~\cite{MNIST} 
& NA\textsuperscript{3}&NA & NA&NA & NA&NA & NA&NA & 6&10 & 0&0 & 4&33 & 153&+ \\

Automlp~\cite{chen2023automlp} 
& 0&1 & 0&0 & 0&0 & 7&5 & 0&1 & 0&0 & 0&1 & 4&5 \\

Particle Tracking~\cite{abidi2022charged} 
& 0&1 & 0&0 & 1&1 & 24&24 & 1&1 & 0&0 & 1&1 & 8&15 \\
\hline
\hline
Custom 1 
& 0&1 & 0&0 & 1&0 & 47&39 & 2&1 & 0&0 & 2&2 & 16&21 \\

Custom 2 
& 0&1 & 2&1 & 2&2 & 98&77 & 5&2 & 2&1 & 4&4 & 54&45 \\

Custom 3 
& 0&1 & 0&0 & 1&1 & 60&52 & 3&1 & 0&0 & 2&3 & 23&36 \\
\hline
\end{tabular}
\end{center}
\raggedright \textsuperscript{1} {\scriptsize  Ground truth} \\
\textsuperscript{2} {\scriptsize  Prediction} \\
\textsuperscript{3} {\scriptsize  Not Applicable}
\par
\label{zcu(g,p)}
\end{table}

\begin{figure}[t]
\centering
\includegraphics[width=\textwidth]{ 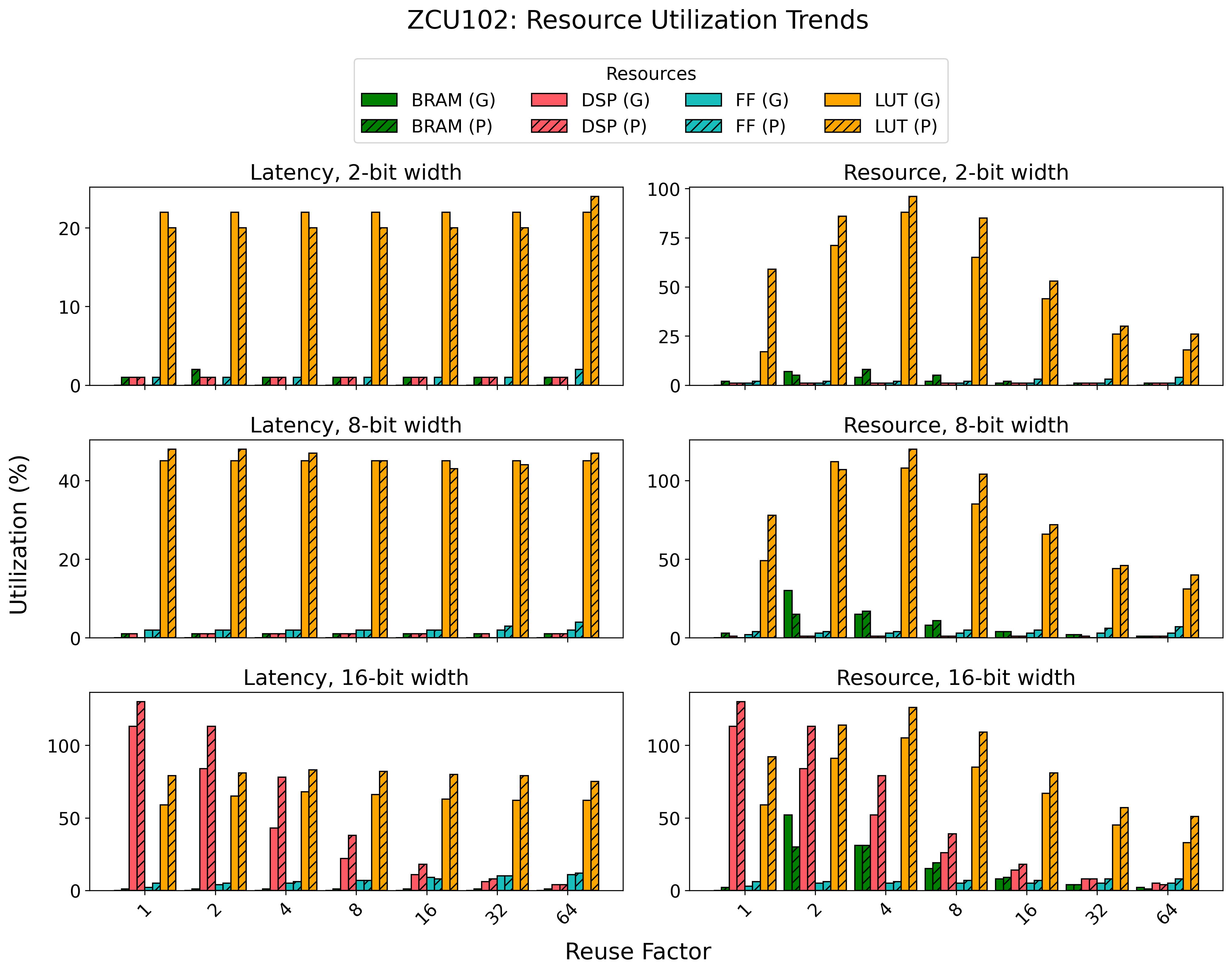}
\caption{Comparing ground truth (G) and prediction (P) trends of resource utilization across different synthesis parameters on the ZCU102. The values are averaged across the benchmark NNs.}
\label{fig:zcu-prediction-trends}
\end{figure}

\begin{figure}[t]
\centering
\includegraphics[width=\textwidth]{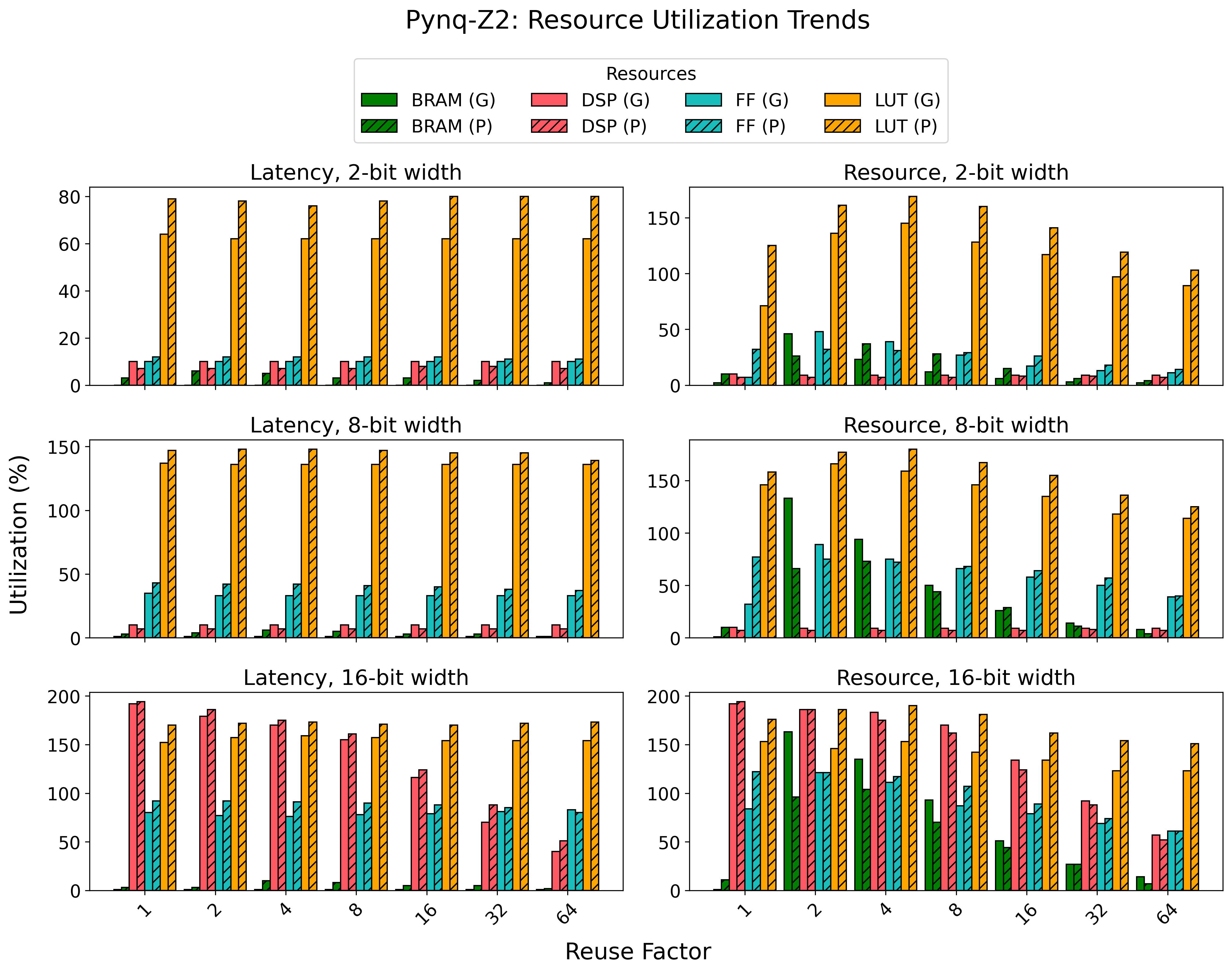}
\caption{Comparing ground truth (G) and prediction (P) trends of resource utilization across different synthesis parameters on the Pynq-Z2. The values are averaged across the benchmark NNs.}
\label{fig:pynq-prediction-trends}
\end{figure}

\begin{figure}[t]
\centering
\includegraphics[width=\textwidth]{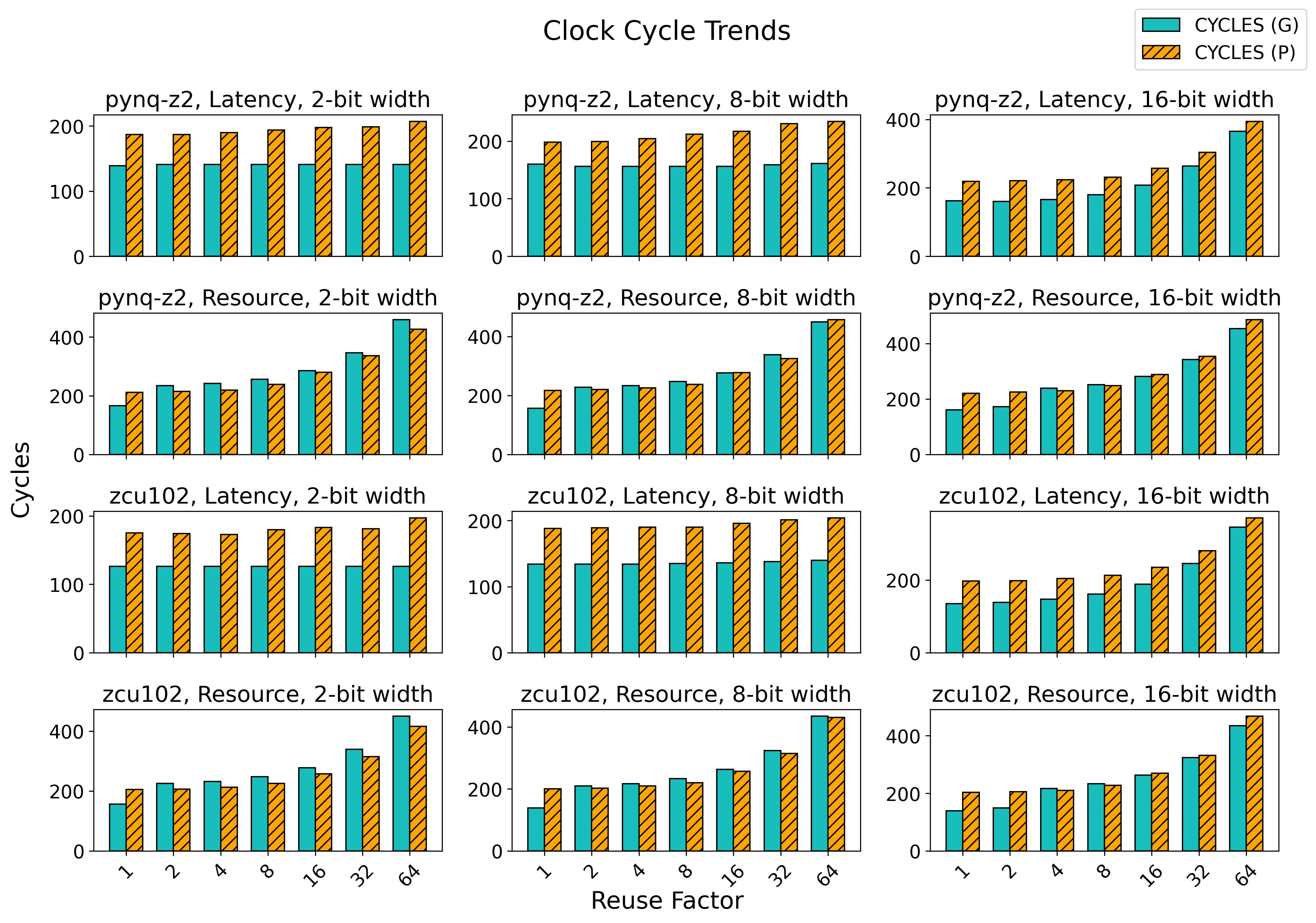}
\caption{Comparing ground truth (G) and prediction (P) trends of clock cycles across different synthesis parameters on the Pynq-Z2 and ZCU102. The values are averaged across the benchmark NNs.}
\label{fig:latency-trends}
\end{figure}

Tables~\ref{zcu(g,p)}~and~\ref{pynq(g,p)} compare ground truth and predicted values of resource utilization on the benchmarks as synthesized on a ZCU102 and a Pynq-z2. Due to the tables' limited space, we standardized the reuse factor to~32 and set the precision to 8-bit width. We also covered both strategies (Latency and Resource). A more comprehensive coverage of other configurations is shown in~\cref{fig:zcu-prediction-trends,fig:pynq-prediction-trends,fig:latency-trends}. These tables do not include the Alveo-U200 results, being in the same family as the ZCU102 board, with the synthesis results for both boards being mostly similar.\\

\begin{table}[t]
\footnotesize
\caption{Comparing ground truth (G) and prediction (P) resource utilization percentage for different benchmarks on Pynq-Z2.}
\renewcommand{\arraystretch}{1.3}
\setlength\tabcolsep{3.65pt}
\begin{center}
\begin{tabular}{c|cc|cc|cc|cc|cc|cc|cc|cc} \hline
\multirow{3}{*}{Benchmark} & \multicolumn{8}{c|}{Latency} & \multicolumn{8}{c}{Resource} \\ 
\cline{2-17} 
                               & \multicolumn{2}{c|}{BRAM}   & \multicolumn{2}{c|}{DSP}    & \multicolumn{2}{c|}{FF} & \multicolumn{2}{c|}{LUT}    & \multicolumn{2}{c|}{BRAM}   & \multicolumn{2}{c|}{DSP}   & \multicolumn{2}{c|}{FF}   & \multicolumn{2}{c}{LUT}  \\  
\cline{2-17} 
                               & \multicolumn{1}{c}{G\textsuperscript{1}} & \multicolumn{1}{c|}{P\textsuperscript{2}} & \multicolumn{1}{c}{G} & \multicolumn{1}{c|}{P}& \multicolumn{1}{c}{G} & \multicolumn{1}{c|}{P}& \multicolumn{1}{c}{G} & \multicolumn{1}{c|}{P}& \multicolumn{1}{c}{G} & \multicolumn{1}{c|}{P}& \multicolumn{1}{c}{G} & \multicolumn{1}{c|}{P}& \multicolumn{1}{c}{G} & \multicolumn{1}{c|}{P}& \multicolumn{1}{c}{G} & \multicolumn{1}{c}{P}  \\ 
\hline \hline
JET~\cite{fahim2021hls4ml} 
& 1&3 & 2&2 & 12&13 & 91&95 & 9&5 & 2&2 & 8&11 & 49&72 \\

Top Quarks~\cite{duarte2018fast} 
& 0&1 & 0&0 & 3&5 & 23&39 & 0&3 & 0&0 & 8&5 & 28&52 \\

Anomaly~\cite{borras2022open} 
& 3&2 & 58&57 & 150&184 & +\textsuperscript{4}&197 & 9&5 & 58&57 & 156&183 & +&199 \\

BiPC~\cite{rahali2024efficient} & 0&0 & 0&0 & 29&26 & +&186 & 17&3 & 0&0 & 18&21 & 131&158 \\

CookieBox~\cite{gouin2022data} 
& 1&2 & 2&2 & 15&15 & 169&175 & 10&5 & 2&2 & 80&160 & +&197 \\

MNIST~\cite{MNIST} & NA\textsuperscript{3}&NA & NA&NA & NA&NA & NA&NA & 38&65 & 5&0 & 197&134 & +&+ \\

Automlp~\cite{chen2023automlp} 
& 1&1 & 1&2 & 4&4 & 25&43 & 2&3 & 1&2 & 4&4 & 24&49 \\

Particle Tracking~\cite{abidi2022charged} 
& 1&3 & 1&1 & 11&12 & 85&89 & 8&5 & 1&1 & 8&11 & 48&70\\
\hline
\hline
Custom 1 
& 1&5 & 5&2 & 21&18 & 164&175 & 16&9 & 5&1 & 14&17 & 87&132 \\

Custom 2 
& 1&6 & 23&13 & 58&46 & +&195 & 30&13 & 23&13 & 47&50 & +&197 \\

Custom 3
& 1&2 & 5&2 & 25&28 & +&197 & 19&5 & 5&2 & 16&27 & 126&173 \\
\hline

\end{tabular}
\end{center}
\raggedright \textsuperscript{1} {\scriptsize  Ground Truth} \\
\textsuperscript{2} {\scriptsize  Prediction}  \\ 
\textsuperscript{3} {\scriptsize  Not Applicable} \\ 
\textsuperscript{4} {\scriptsize  Above 200\% Utilization}
\par
\label{pynq(g,p)}
\end{table}

In both tables, the trained models accurately predict resource utilization overall. Looking closely, BRAM and DSP predictions usually perform better than LUT and FF. The LUT prediction remains consistently accurate and occasionally aligns perfectly with the ground truth for certain networks. Furthermore, the FF prediction shows more errors than the other resources.\\

The "NA" in the table indicates that HLS4ML fails to synthesize the corresponding configuration. This failure occurs due to a Vivado limit on the maximum number of parameters per unrolled layer, which cannot exceed 4096. Consequently, we were unable to generate "MNIST" ground truth values and predictions for the latency strategy.\\

Upon comparing both tables, it is clear that due to the lower available resources of Pynq-Z2, Table~\ref{pynq(g,p)} generally exhibits higher resource utilization percentages. Notably, utilization exceeds 200\% for some networks, which we denote with a "+" symbol. We consider 200\% a reasonable cutoff limit, as any NN surpassing this threshold would undoubtedly exceed the board's capacity, leading to timing issues at the implementation phase. Finally, we observed that employing the resource strategy typically leads to a significantly higher BRAM usage especially for Pynq-Z2, but lower LUT, a trend that the trained model adeptly predicts.\\

The predictors should demonstrate the same trends as the ground truth when changing configuration values. Specifically, the precision and reuse factor were incrementally changed to compare the trends, shown in Figure~\ref{fig:zcu-prediction-trends} for ZCU102 and in Figure~\ref{fig:pynq-prediction-trends} for Pynq-Z2 board. The values are averaged across the benchmarks. The prediction trend follows the ground truth trend well for all variables. Generally, a higher precision increases resource usage, while a higher reuse factor reduces DSP utilization. In addition, the predictors tend to slightly overestimate resources.\\

In the next test, we use the same previous process, this time evaluating the trained latency predictor. Table~\ref{latency} compares ground truth and prediction for latency across benchmarks on both ZCU102 and Pynq-Z2. Similar to the resource prediction tables, we have standardized reuse and precision to 32 and 8-bit width respectively, due to space limitations. A comprehensive coverage of other configurations is provided through the subsequent figures.\\

According to Table~\ref{latency} and as expected, we observe that the latency of the resource strategy is typically higher than that of the latency strategy. Moreover, the latency of the benchmarks on the Pynq-Z2 board is generally higher than the ZCU102. Overall, predictions for all cases are generally accurate, with an average error lower than 100 clock cycles. Given that we used a~\SI{10}{ns} clock period for synthesis, this translates to an average latency difference lower than 1~\si{\micro\second} between prediction and ground truth. Similar to the previous tables, we could not provide ground truths for the "MNIST" benchmark with the latency strategy.\\

Finally, the trend of latency prediction versus ground truth is presented in Figure~\ref{fig:latency-trends}, by incrementally adjusting precision and reuse factor. The latency values are averaged across the benchmark architectures. The figure shows that increasing precision and reuse factors lead to higher latency, a trend predicted accurately by the trained models. Similarly to the resource prediction results, errors primarily occur as overestimation rather than underestimation.

\begin{table}[t]
\footnotesize
\caption{Comparing the latency (number of clock cycles) between ground truth (G) and prediction (P) across various benchmarks on both ZCU102 and Pynq-Z2 boards.}
\renewcommand{\arraystretch}{1.4}
\begin{center}
\begin{tabular}{c|cc|cc|cc|cc}
\hline
\multirow{2}{*}{Benchmarks} & \multicolumn{4}{c|}{Latency} & \multicolumn{4}{c}{Resource} \\ \cline{2-9}
& \multicolumn{2}{c|}{ZCU102} & \multicolumn{2}{c|}{Pynq-Z2} & \multicolumn{2}{c|}{ZCU102} & \multicolumn{2}{c}{Pynq-Z2} \\\hline 
& G\textsuperscript{1} & P\textsuperscript{2} & G & P & G & P & G & P \\ \hline \hline
JET~\cite{fahim2021hls4ml} & 37&57 & 52&83 & 167&180 & 177&188 \\
Top Quarks~\cite{duarte2018fast} & 19&44 & 26&70 & 16&159 & 24&149 \\
Anomaly~\cite{borras2022open} & 296&333 & 334&399 & 459&356 & 492&436 \\
BiPC~\cite{rahali2024efficient} & 85&92 & 104&98 & 280&166 & 289&177 \\
CookieBox~\cite{gouin2022data} & 535&557 & 549&583 & 663&691 & 674&678 \\
MNIST~\cite{MNIST} & NA&NA & NA&NA & 871&702 & 884&710 \\
Automlp~\cite{chen2023automlp} & 27&36 & 64&61 & 134&161 & 144&163 \\
Particle Tracking~\cite{abidi2022charged} & 34&56 & 49&84 & 159&189 & 169&180 \\ \hline \hline
Custom  1 & 45&66 & 68&95 & 211&249 & 225&249 \\
Custom  2 & 209&232 & 245&268 & 410&368 & 435&416 \\
Custom  3 & 94&94 & 119&123 & 191&222 & 203&226 \\ \hline
\end{tabular}
\end{center}
\hspace*{0.125\textwidth} \textsuperscript{1} {\scriptsize  Ground truth} \\
\hspace*{0.125\textwidth} \textsuperscript{2} {\scriptsize  Prediction}
\par
\label{latency}
\end{table}

\section{Discussion}

For resource prediction, we consider an error lower than 10\% to be acceptable. According to predictions on the synthetic test set (Figure~\ref{fig:boxplot-merged}), we found that, on average, 81\% of resource predictions fall within this acceptable range. Breaking it down by resource type, 80\% of BRAM predictions, 89\% of DSP predictions, 88\% of FF predictions, and 66\% of LUT predictions are within the 10\% error threshold. Our results reveal that the LUT prediction errors are generally higher than the rest of the resources (Figure \ref{fig:boxplot-merged}). This is primarily because LUT is the most used resource, with its ground truth covering a wide range of values. For example, even a configuration with low precision and high reuse that shows negligible usage of BRAM, DSP, and FF can still employ a significant amount of the FPGA's LUTs. In contrast, the furthest outliers of the FF errors indicate that the corresponding model might struggle to make accurate FF predictions. We attribute this to two reasons: the distribution of FF values in the dataset is heavily skewed, where most values are close to the lower and upper thresholds (0\% and 200\%). However, this is also the case for BRAM and DSP. More importantly, it is difficult to determine the factors that affect FF usage the most in NN implementation on FPGA, as evidenced by the low correlation with FF in Figure~\ref{fig:corr-matrix}. \\

The predictive performance for clock cycles is similarly commendable. Despite the broader IQR, reflecting higher variability in the predictions due to the inherent complexity of latency measurement, 85\% of prediction errors are below 100 clock cycles. Considering a clock frequency of~\SI{10}{n\second}, this results in an error of less than~\SI{1}{\micro{\second}} for those predictions, which we find satisfactory.\\

Since all the tests were based on Fully Connected Neural Networks (FCNNs), we observed minimal DSP and BRAM utilization across most models, particularly on the larger ZCU102 board. We also noticed a substantial increase in BRAM utilization with the resource strategy. Generally, the implementation of fully connected layers is optimized to use more LUTs, and fewer DSPs, fitting even smaller boards. However, this is not the case for other popular architectures, which are worth exploring. Based on previous work~\cite{rahimifar2023exploring}, we anticipate that CNNs behave differently, exhibiting higher DSP usage across all boards. Additionally, even the largest FCNNs had a latency lower than \SI{100}{\micro\second}, and deep CNNs are expected to occupy a higher range.\\

As mentioned earlier, all the training sets and test sets were generated targeting a ~\SI{10}{n\second} clock period. While the clock cycle is directly related to the latency time, we also observed that changing the target clock period affects resource utilization. For instance, decreasing it from \SI{10}{n\second} to \SI{5}{n\second} heavily increases FFs utilization and only slightly changes LUTs. Accordingly, future iterations will aim to generate a larger dataset with a diverse range of clock periods in order to predict these trends.\\

Generating data and synthesizing a vast array of NNs for our training set was time-consuming. However, we consistently noticed improvements in our predictions as the dataset grew larger and incorporated more configurations. The data generation space for this task is significant, and we believe there is a massive potential in exploring even wider hyperparameter combinations and HLS4ML configurations. For instance, the synthetic data can incorporate complex architectures with other layer types, increased layer sizes, additional activation functions, and explore different precision and higher reuse values. We believe these efforts will lead to a robust tool, capable of handling the majority of architectures found in literature, and we anticipate more accurate predictions as the dataset expands. Collecting a larger dataset with more architectures (FCNN, CNN, RNN) will be part of our future work and will require significantly more CPU time than the current dataset.\\

It is worth mentioning that some networks' synthesis showed unexpected behaviors, which has undoubtedly impacted the learning process of predictive models. For instance, theoretically, increasing the reuse factor should lead to a reduction in DSP usage with a low impact on LUT and an increase in latency. However, we observed instances where this was not the case, with reports displaying peculiar behavior (Resources and latency results with 2-bit and 8-bit precisions in Figures.~\ref{fig:zcu-prediction-trends}, and~\ref{fig:pynq-prediction-trends}). While the number of such NNs is not significant in the training set, we still find this worth mentioning since it could potentially affect the predictors' training and introduce further errors.\\

At this stage, our focus has been solely on the synthesis report generated by Vivado HLS, rather than the Vivado implementation reports. Through extensive testing of various NN implementations, we observed that the Vivado HLS report serves as a preliminary assessment, subject to further Vivado optimization during the implementation phase. This observation is particularly applicable to LUTs, often overestimated by Vivado HLS, with actual utilization decreasing post-implementation. While we believe that overestimation is preferable to underestimation in this context, we intend to experiment with the implementation reports in the future and address the effects of Vivado optimizations. This will provide users not only with an approximation of the synthesis report but also an insight into what to expect in the final implementation report.\\

\section{Conclusion}

Implementing ML models on FPGAs is becoming common across various domains, leveraging FPGA's reduced latency compared to alternative processing units for accelerating ML inference tasks. However, developing NNs for FPGAs is significantly time-consuming. HLS4ML accelerates this process but lacks the ability to inform users if an NN fits the targeted board before synthesis. This is problematic as larger networks can take hours to synthesize, with resource utilization and latency reports available only after a successful synthesis. Each failed synthesis further slows down the development process.\\

This paper demonstrates the efficiency of using ML models to predict both resource utilization and latency reports for NNs prior to FPGA implementation. The prediction models, tested on both generated NNs and existing benchmark architectures, showed high accuracy, achieving low average errors and high R\textsuperscript{2} scores across various dependent variables. While our results reveal commendable prediction accuracy and minimal error for resources and latency, our current scope is confined to predicting FCNNs. Considering the time required to generate a larger dataset with other architectures, we limited ourselves to FCNNs as a proof of concept, planning to extend our work to cover CNN and RNN architectures. Additional work will take place to improve the prediction accuracy as we move forward. \\

Ultimately, we believe that this work holds potential beyond resource and latency prediction. By leveraging these early predictions, it is possible to develop an optimization tool that can assist users in preparing their models for the targeted board early in development. This approach promises significant time savings, particularly for individuals lacking expertise in hardware architecture. With the increasing adoption of ML accelerators on FPGA platforms, many software developers with limited hardware knowledge seek to implement their models on FPGA. The proposed tool stands as a notable addition to existing ML-to-FPGA tool-flows, aiming to greatly aid users in navigating the conversion process effectively.

\section*{Funding}

This research was undertaken, in part, thanks to funding from the Canada Research Chairs Program. A. C. Therrien holds the Canada Research Chair in Real-Time Intelligence Embedded for High-Speed Sensors.

\section*{Disclosures}

The authors declare no conflicts of interest. 

\section*{Ethics statement}

This paper reflects the authors' research and analysis in a truthful and complete manner.

\section*{Data availability}

The data that support the findings of this study is publicly available.

\section*{Acknowledgments}

Each author contributed equally to each aspect of this study, including research design, data analysis, and manuscript preparation. The authorship order does not signify importance but reflects our collaboration. We extend our gratitude to the HLS4ML team for their open discussions, which helped our work. We also thank our colleague, Charles-Étienne Granger for providing us additional computational resources to accelerate the data generation task.

\section*{References}
\bibliographystyle{unsrt} 
\bibliography{References} 

\end{document}